# Advancing 3D finger knuckle recognition via deep feature learning

Kevin H. M. Cheng[1], Xu Cheng[1,2], and Guoying Zhao[1*]

[1]Center for Machine Vision and Signal Analysis, University of Oulu, Finland
[2]Nanjing University of Information Science and Technology, China

[*]Corresponding author. Email: guoying.zhao@oulu.fi;
Contributing authors: homan.cheng@oulu.fi; xcheng@nuist.edu.cn;

**Abstract.** Contactless 3D finger knuckle patterns have emerged as an effective biometric identifier due to its discriminativeness, visibility from a distance, and convenience. Recent research has developed a deep feature collaboration network which simultaneously incorporates intermediate features from deep neural networks with multiple scales. However, this approach results in a large feature dimension, and the trained classification layer is required for comparing probe samples, which limits the introduction of new classes. This paper advances this approach by investigating the possibility of learning a discriminative feature vector with the least possible dimension for representing 3D finger knuckle images. Experimental results are presented using a publicly available 3D finger knuckle images database with comparisons to popular deep learning architectures and the state-of-the-art 3D finger knuckle recognition methods. The proposed approach offers outperforming results in classification and identification tasks under the more practical feature comparison scenario, i.e., using the extracted deep feature instead of the trained classification layer for comparing probe samples. More importantly, this approach can offer 99% reduction in the size of feature templates, which is highly attractive for deploying biometric systems in the real world. Experiments are also performed using other two public biometric databases with similar patterns to ascertain the effectiveness and generalizability of our proposed approach.

## 1 Introduction

Physiological biometrics such as face [1-2] , iris [3-4], ear [5-5], fingerprint [7-8], palmprint [9-10], and vein patterns [11-13] can provide reliable and convenient solutions to automatically recognize human identities in real world applications such as authentication for immigration inspection, unlocking personal devices and approving financial transactions. Finger knuckle pattern [14-16], the skin crease and valley pattern between the middle and proximal phalanges of fingers, contains discriminative information, which is visible from a distance, and can be conveniently imaged simultaneously with other hand biometrics. It is an emerging alternative biometric identifier which can be used alone or together with other identifiers for multi-modal biometric recognition [17-19].

With the advancement of 3D reconstruction techniques [20-21], the investigation of biometric recognition using 3D information has become a popular research trend [22-23]. While finger knuckle patterns are essentially 3D in nature, the use of such 3D information offers complimentary and illumination invariant information for more accurate and robust biometric recognition [24-26]. Deep learning technologies have been widely investigated in many computer vision tasks including biometrics [27-28]. State-of-the-art development of 3D finger knuckle recognition also adopted a deep learning based approach [29]. This method simultaneously incorporates the intermediate features from deep neural networks with multiple scales to accommodate the irregular finger knuckle patterns. However, the size of feature templates is of one eighth of the input resolution (i.e., 6×10) and a concatenation of intermediate features result in 832 channels, which generates a large feature dimension of 49920. Another inherit limitation of this method is that the trained classification layer is required for comparing probe samples with the gallery samples, which limits the introduction of unseen subjects/classes.

This paper addresses the above limitations by incorporating the advantages of both conventional deep learning approaches and the 3D finger knuckle recognition approach (FKNet), with the objective of investigating the possibility of learning a discriminative feature vector with minimum dimension for representing each 3D finger knuckle image. The key contributions of this paper can be summarized as follows:

(i) Conventional deep learning approaches generate abstract feature vectors without considering the alignment of spatial features while a state-of-the-art deep learning based 3D finger knuckle recognition approach, FKNet [29] incorporated multi-scale features and with alignment. However, the former approach is limited by large variation between the training and test samples, especially for the 3D finger knuckle recognition; the latter approach is limited by its bulky feature size. This paper firstly investigates a more practical feature comparison scenario, i.e., comparing the similarities between the extracted features without using the trained classification layer, and exposes a inherit limitation of FKNet which simultaneously incorporates intermediate features with multiple scales and constitutes a large feature dimension. This important finding encourages the development of more compact feature description methods meanwhile preserving the discriminative spatial features for 3D finger knuckle recognition.

(ii) This paper incorporates the advantages of both conventional deep learning approaches and the 3D finger knuckle recognition approach (FKNet) by developing a new approach to encode the discriminative spatial features with a minimum feature size. The network backbone can be inherited from FKNet, but we replace the bulky classification layer to a feature extraction layer, which enables flexible feature dimensions. Those encoded feature vectors from probe samples can be easily compared with gallery samples. From our reproducible experimental results [30] presented in Section 4, we achieve state-of-the-art recognition performances with a huge reduction in feature size (e.g., from 49920 to 600), which is highly desirable for deploying biometric systems in the real world. More importantly, we further analyze the recognition performance using both the classification scenario and the feature comparison scenario. We show that our approach significantly outperforms the original approach in the more practical feature comparison scenario.

(iii) Lastly, this paper provides open-source implementation codes [30] using a recent version of PyTorch framework (1.11.0), which can help to further advance research on 3D finger knuckle recognition.

Table 1 summarizes the key performances between the proposed approach and the existing state-of-the-art method. The numerical figures are extracted from the experimental results presented in Section 4.

**Table 1.** Comparative summary of the proposed approach with a state-of-the-art method. Better performances are in bold.

|  | Feature Size | Classification Scenario | | Feature Comparison Scenario | |
|---|---|---|---|---|---|
|  |  | EER ↓ | Rank1 ↑ | EER ↓ | Rank1 ↑ |
| FKNet [29] | depends on input size, e.g., 49920 | **5.5%** | 91.1% | 16.6% | 65.5% |
| This Paper | flexible, e.g., **600, 800** | 5.8% | **91.9%** | **4.7%** | **92.5%** |

## 2   Related Work

Finger knuckle biometrics has been well summarized in recent publications [31-32]. One of the earliest works [24] investigated the use of finger knuckle prints as a biometric identifier. After that, there are several investigations and advancements on finger knuckle recognition [33-35], while more recent methods include a holistic approach [36], a LPQ approach [37] and a deep learning approach [38]. The study can also be extended to a multi-modal fusion with other hand biometrics [39-40] and 3D domain [26, 41].

Since the first systematic study on 3D finger knuckle recognition [26], research efforts have been made on the development of more efficient matching scheme [42], more accurate feature representations with similarity functions [43-44], and specialized deep neural networks [29]. Although 3D finger knuckle recognition itself is a task of image recognition, it has been shown in reference [29] that the performances of generic deep neural networks such as ResNet [45] suffer from limited training data and the large variance between the distributions of the train and test dataset. These challenges limit the capabilities of generic neural networks to generalize the classification of true identities. Therefore, investigating an effective neural network architecture or a learning approach for the challenging 3D finger knuckle recognition task is a much needed research work. This paper extends the state-of-the-art deep learning based 3D finger knuckle recognition framework [29] to accommodate more practical feature comparison scenarios for the development of real world biometrics applications.

## 3   Methodology

### 3.1   Data Preprocessing

We adopt the same preprocessing steps as in FKNet [29]. Raw photometric stereo based finger knuckle images are segmented using Mask R-CNN [46] where ground truth masks of the training samples are first automatically computed using the edge pixel counting method described in reference [26], then the outlining samples are manually discarded. This network is fine-tuned from the pre-trained weights on COCO dataset [47]. The 3D finger knuckle images are reconstructed using a conventional photometric stereo method [48]. These images are then rendered to the illumination invariant images as described in reference [29]. The rendered images are then fed to the neural network presented in Section 3.2.

### 3.2   Finger Knuckle Network Plus

Inspired by FKNet [29], beside the final output of neural networks, intermediate deep features with different spatial scales also embed useful information for distinguishing finger knuckle patterns. Furthermore, translational alignment is a key successful component to accommodate a large degree of irregular translational variation of the finger knuckle patterns. With these techniques, FKNet significantly outperforms other generic neural networks in the challenging experimental settings that, (i) very limited data used for the training (only one raw image for each of the 190 subjects); (ii) large variance between train and test dataset, which are acquired under two different imaging lenses (can be observed in Figure 1); (iii) large translational variations due to undefined finger knuckle pattern boundaries. In fact, the samples acquired from the same lens are similar to each other and very different from those acquired from a different lens. FKNet utilized more deep features in intermediate layers meanwhile retaining the spatial information to an extent. This approach is inherently limited by having large feature size and the trained classification layer is needed to process the large feature space into correct prediction of classes. Therefore, we connect the useful mechanism from FKNet with generic neural networks approach which output a dimensionally reduced feature vector for each sample. We incorporate feature collaboration, input alignment and dimension reduction to develop the proposed FKNet+.

We adopt the backbone from FKNet which is modified based on ResNet-50 [45]. Figure 1 shows the overall architecture of the proposed FKNet+ and the details will be presented as follows. The layers refer to the residual layers. A building block of the residual structure can be defined as:

$$y = f(x, w, b) + x \tag{1}$$

where $x$ is the input of the layer; $y$ is the output; $w$ is the weights; $b$ is the bias; and $f$ is a generic function. The network is then constructed with:

$$M = f_3\big(f_2(f_1(I))\big) + g_2\big(f_2(f_1(I))\big) + g_1(f_1(I)) \tag{2}$$

where $I$ is the input image; $f_1$, $f_2$, and $f_3$ are the functions corresponding to the convolution layers; $g_1$ and $g_2$ are the functions corresponding to the max pooling layers so that the deep features resized to the same spatial dimension can be concatenated; and $M$ is the output feature of FKNet. Unlike FKNet which directly passes the large feature $M$ to a classification layer, we insert a feature extraction layer:

$$v = f_{cls}(f_{fea}(M)) \tag{3}$$

where $f_{fea}$ is a feature extraction layer consisting of a convolution layer, then an activation layer (ReLU) and then a normalization layer (BatchNorm). Finally, the dimensionally reduced feature vector $v$ (e.g., 1×1×600) is passed to a classification layer $f_{cls}$ with cross entropy loss for the training. Since the alignment scheme in FKNet requires a fully convolutional architecture, we also use convolution layers instead of fully connected layers in both our feature extraction layer $f_{fea}$ and the classification layer $f_{cls}$. For the training in open-set scenario, we replace the training of using single branched classification layer by using a Siamese architecture with cosine embedding loss between first and second session's training data.

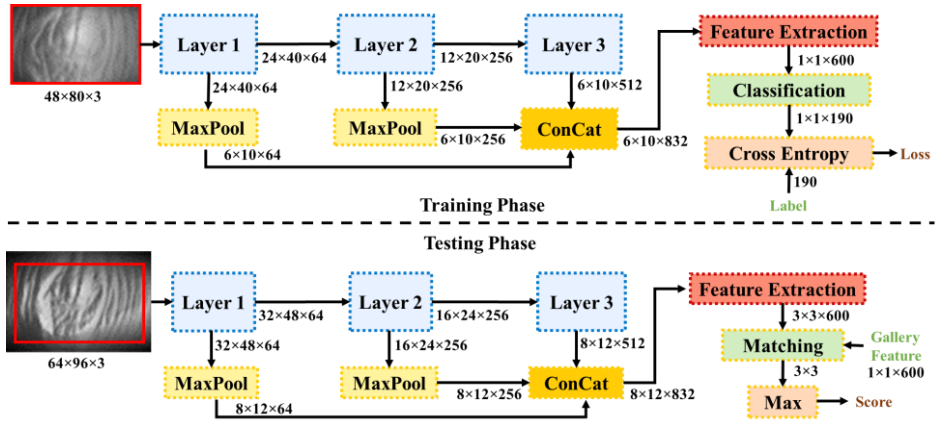

**Fig. 1.** The network architecture for the proposed FKNet+. The finger knuckle images are extracted from the preprocessed dataset [29].

**Table 2.** Layer configurations of FKNet+ during the training.

| Layer | Output Size | Kernel Size |
|---|---|---|
| Layer 1 | $24 \times 40 \times 64$ | $7 \times 7, 64$, stride 2 |
| Layer 2 | $12 \times 20 \times 256$ | $3 \times 3$ max pool, stride 2 <br> $\begin{bmatrix} 1 \times 1, 64 \\ 3 \times 3, 64 \\ 1 \times 1, 256 \end{bmatrix} \times 3$ |
| Layer 3 | $6 \times 10 \times 512$ | $\begin{bmatrix} 1 \times 1, 128 \\ 3 \times 3, 128 \\ 1 \times 1, 512 \end{bmatrix} \times 4$ |
| Feature Extraction | $1 \times 1 \times 600$ | $6 \times 10, 600$ |
| Classification | $1 \times 1 \times 190$ | $1 \times 1, 190$, *softmax* |

### 3.3 Evaluations with Finger Knuckle Network Plus

During the testing, a softmax layer is used instead of the classification layer. Each image with a larger spatial resolution (e.g., 64×96 vs 48×80) can be inputted to the network, resulted in multiple (e.g., 3×3) feature vectors v and therefore multiple classification prediction scores. The entire probability vector of 190 classes containing the maximum probability value among the matrix will be the final prediction scores.

Our approach can also be easily used in a feature comparison scenario, i.e., comparing the test samples with stored feature templates. In this situation, the extracted feature vectors can be compared using mean square error or cosine similarity. In the case of the input alignment, each multiple feature vectors resulted from the input of an image with a larger spatial resolution will be compared with the gallery feature vectors, which is the extracted feature vectors from the train set. The minimum scores of the mean square error or cosine similarity computation will be the final comparison scores. For the open-set scenario, we compute the cosine similarities between the embeddings from first and second session's test data. Table 2 shows the layer configurations with an input image of size 48×80×3 and a total number of 190 classes.

## 4 Experimental Analysis and Results

This section presents the comparative experimental results to investigate the effectiveness of various configurations of the existing and proposed methods, and to validate the experimental performances of the proposed approach. The experiments are performed on a system with Ubuntu 18.04 with PyTorch 1.11.0, while the hyperparameters can be found from our open-source implementation codes [30]. In the performance evaluation, receiver operating characteristics (ROC) curves with equal error rates (EER) and cumulative match characteristics (CMC) curves with rank one accuracies are used as the evaluation metrics. The data that support the findings of this study are available from respective references of databases cited in Section 4.1-4.3.

### 4.1 3D Finger Knuckle Recognition

The HKPolyU 3D Finger Knuckle Images Database [49] is recently the only publicly available 3D finger knuckle images database. This database provides 2508 forefinger 3D images and 2508 middle finger 3D images acquired from 228 subjects. Among all subjects, 190 subjects are with images acquired in two sessions. Among the 190 subjects, 89 subjects are with images acquired using two different lenses, which provides challenging cross domain images (can be observed from Figure 1). We follow the same experimental protocol as in reference [29] that, only the first images per each of the 190 subjects in session one are used as the train set while all six images per subjects in session two are used as the test set. During training, rotationally shifted versions of ±10 degrees of the training images are used for data augmentation. This evaluation protocol generates 215460 (190×189×6) imposter comparison scores and 1140 (190×6) genuine comparison scores. It is noteworthy that first session's data is considered as the training/gallery set like biometric enrolment or registration while the second session's data is considered as the test or probe set like biometric evaluation. There is no overlapping between the training and test dataset and this protocol is fair and reliable for comparing the performances of methods. In practice, this scenario simulates the deployment in a small-scale environment like personal devices or home, where retraining networks is very easy.

**Ablation Studies**

To investigate the effectiveness of various configurations, we perform two sets of comparative experiments for input alignment and dimension of feature vectors.

Firstly, we examine the performances of four popular generic deep learning architectures, i.e., ResNet-50 [45], DenseNet-121 [50], GoogleNet [51], ViT-b-16 [52], without input alignment. These models are fine-tuned from their respective pre-trained weights, except for ViT-b-16, we modified the input size for adapting the 3D finger knuckle dataset. We modify the last classification layer of the networks to an output of 190 classes for this dataset. Figure 2(a)/(b) shows the experimental results.

FKNet [29] is a state-of-the-art deep learning based method for 3D finger knuckle recognition. This model is fine-tuned from the pre-trained weights of ResNet-50. It is suggested that the input alignment technique, i.e., presenting a combination of eight horizontally shifted and eight vertically shifted versions of the test samples to the trained network while taking the maximum prediction scores among these shifted versions as the final score, is effective for FKNet. Therefore, it is interesting to investigate whether the four generic deep learning methods can also be benefitted with the input alignment technique. Figure 2(c)/(d) shows the experimental results using the input alignment technique. By comparing figure 2(a) with 2(c) and 2(b) with 2(d), it can be observed that the input alignment technique is slightly effective for the four generic deep learning methods as shown in the ROC and CMC curves. It can be noted that the results presented in this paper is not exactly the same as in the original reference of FKNet, because we re-implement the methods using the latest version of PyTorch framework. It can be expected that this framework behaves differently in the initialization, batch normalization, randomness in internal operations and data shuffling. To ensure fairness in comparisons with different methods, we implement and perform experiments using the same system and configurations unless specified. The results can also be fully reproduced from our implementation codes [30].

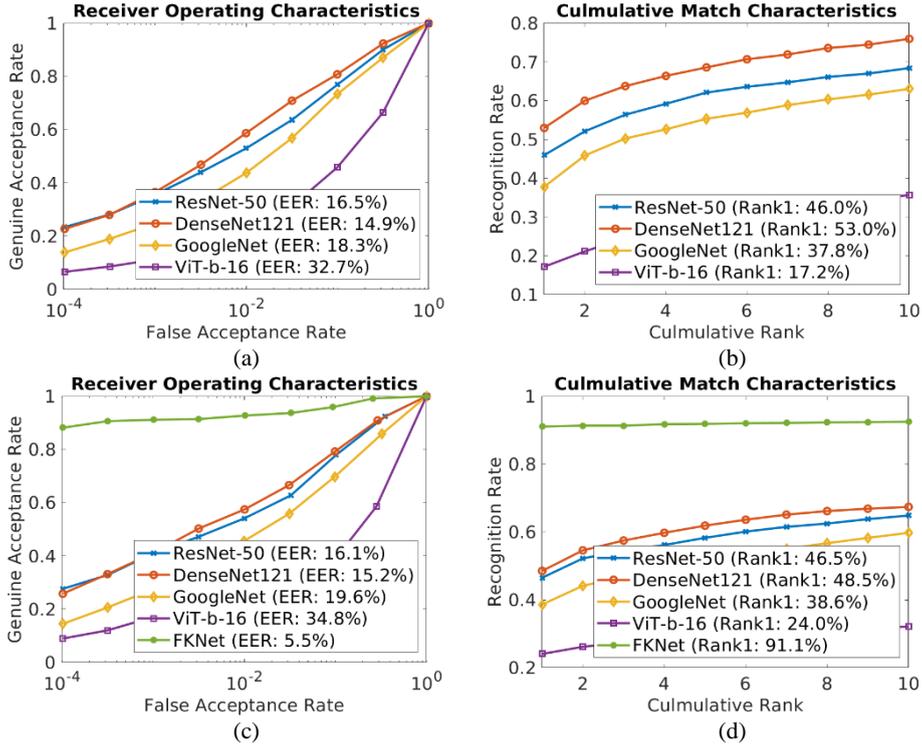

**Fig. 2.** Baseline methods under the classification scenario, ROC and CMC: (a)/(b) without input alignment; (c)/(d): with input alignment.

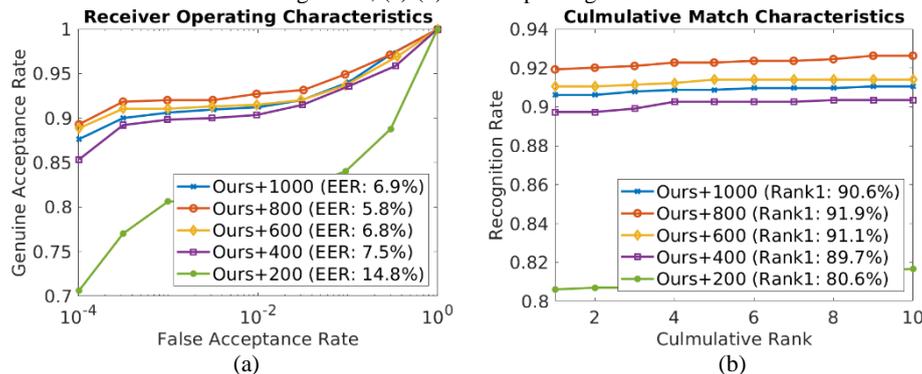

**Fig. 3.** Our method with varying feature sizes under the classification scenario: (a) ROC; (b) CMC.

Secondly, we investigate the minimum feature dimension size required for representing the finger knuckle patterns. The proposed FKNet+ can be with any arbitrary feature dimension size. This model is fine-tuned from the pre-trained weights of ResNet-50. Reference [26] suggested that finger knuckle feature representation can be modeled with 886 independent pixels, it is reasonable to experiment with a feature dimension of 1000, and then decrease. We have evaluated with a feature dimension of 1000, 800, 600, 400 and 200. Figure 3(a)/(b) shows the experimental results using our method with varying feature sizes. It can be observed that using a feature dimension of 800 produces the best performance.

For deploying biometric systems in real world, it is important to allow end users to update their gallery data, i.e., allowing registration of new identities. Therefore, the performance of feature comparison, i.e., comparing the test samples with stored feature templates, is an important evaluation for accessing the flexibility of the methods. Unlike the classification approach which trained a classification layer to predict the probabilities of each class, we evaluate the performance of comparing the deep features, extracted from the layer before the classification layer, between the train and test set. Figure 4(a)/(b) shows the experimental results using our method with varying feature sizes under this feature comparison scenario, using mean square error when comparing the features, while Figure 4(c)/(d) shows such experimental results using cosine similarity when comparing the features. By comparing figure 4(a) with 4(c) and 4(b) with 4(d), it can be observed that the using cosine similarity produces better performances, while using a feature dimension of 600 produces the best performance.

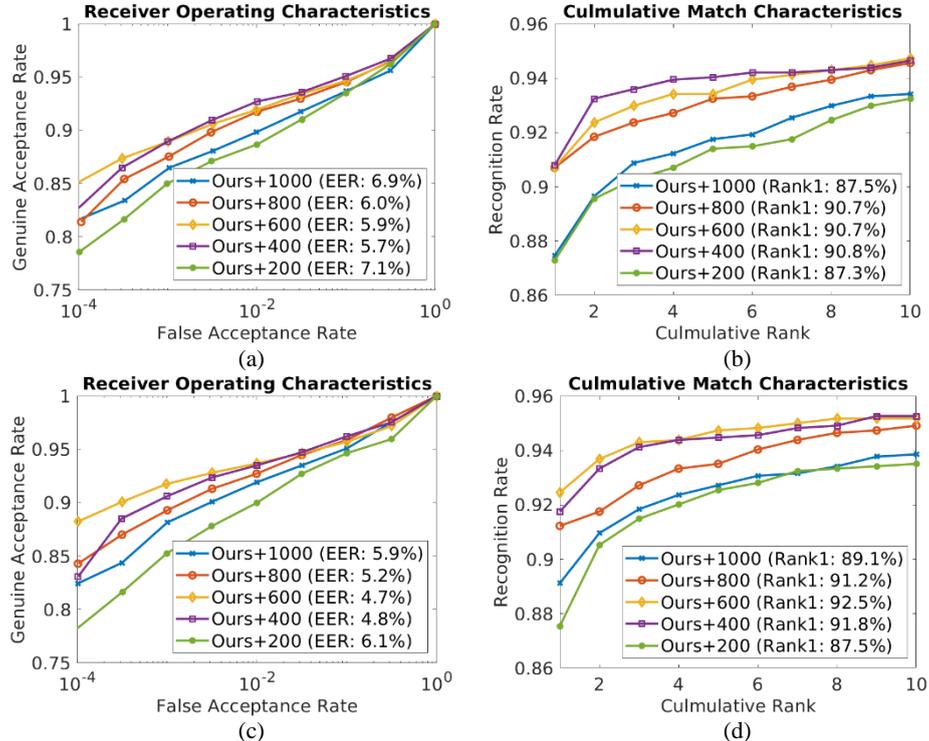

**Fig. 4.** Our method with varying feature sizes under the feature comparison scenario, ROC and CMC: (a)/(b) using mean square error; (c)/(d): using cosine similarity.

**Comparisons with state-of-the-art methods**

From the experimental results of ablation studies, it can be observed that FKNet significantly outperforms the four generic deep learning methods including ResNet-50, DenseNet-121, GoogleNet, and ViT-B-16. Therefore, FKNet is selected as the best performing candidate for performance comparisons. For our proposed FKNet+, we select a feature dimension of 800 which produces the best performance under the classification scenario. Figure 5(a)/(b) shows the comparative experimental results with the state-of-the-art method. It can be observed that our proposed method outperforms the best performing candidate, FKNet. More importantly, the proposed FKNet+ only requires a feature dimension of 800 while FKNet requires a feature dimension of 49920.

For the feature comparison scenario, we further compare with two more state-of-the-art hand-crafted feature based methods for 3D finger knuckle recognition, i.e., the Surface Gradient Derivatives (SGD) method [26] and its more efficient version with the use of Surface Key Point (SKP) [42]. 3D surface normal images are used for the hand crafted feature extractions with these methods. For our proposed FKNet+, we select a feature dimension of 600 and 800 which produce the best performances. Figure 6(a)/(b) shows the comparative experimental results with the state-of-the-art methods under the feature comparison scenario. It can be observed that FKNet significantly suffered from the feature comparison scenario while the proposed FKNet+ can still offer promising verification (in ROC) and identification (in CMC) performances, which achieves a new state-of-the-art. More importantly, the proposed FKNet+ only requires a feature dimension of 600/800 while FKNet requires a feature dimension of 49920, and SGD/SGD+SKP requires a feature dimension of 7000.

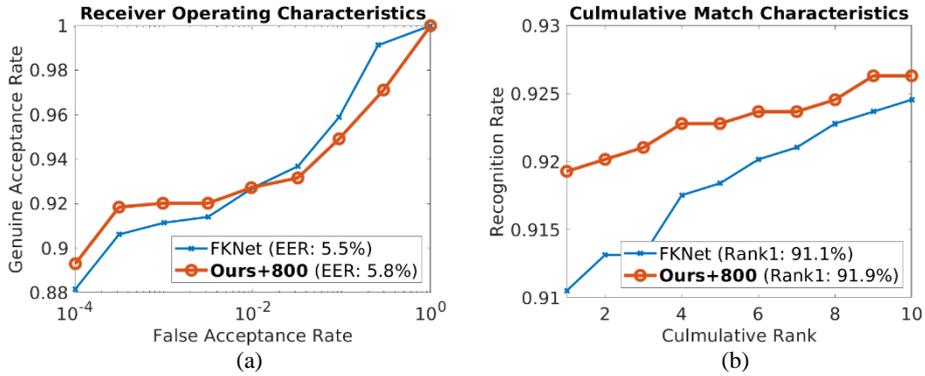

**Fig. 5.** Comparisons with the state-of-the-art method under the classification scenario: (a) ROC; (b) CMC.

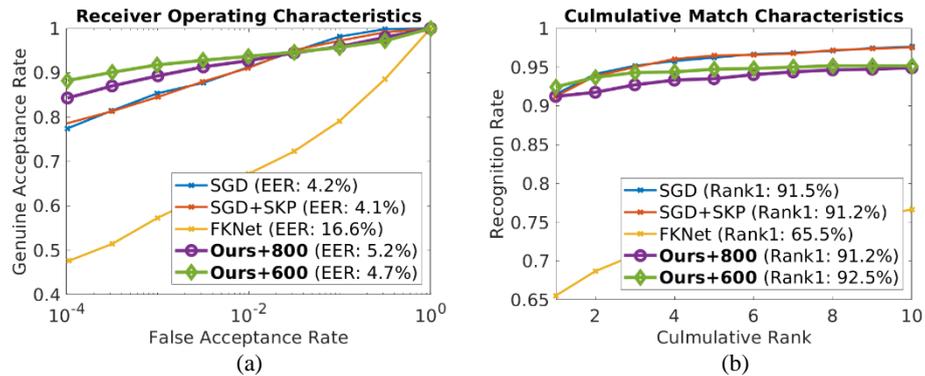

**Fig. 6.** Comparisons with the state-of-the-art methods under the feature comparison scenario: (a) ROC; (b) CMC.

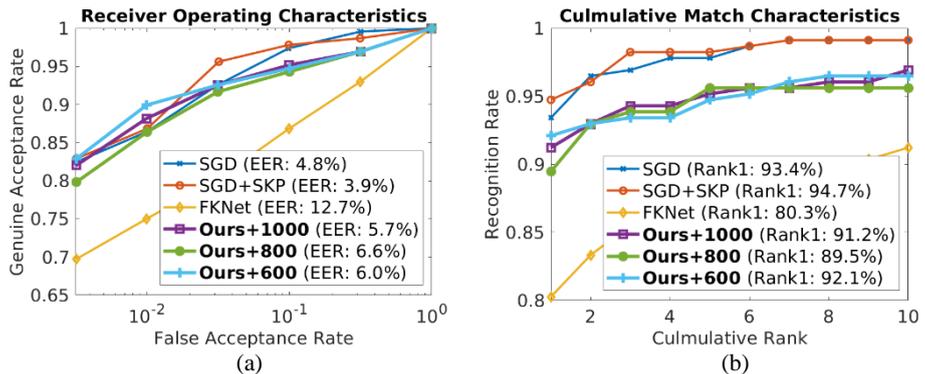

**Fig. 7.** Comparisons with the state-of-the-art methods under the feature comparison scenario with the open-set evaluation protocol: (a) ROC; (b) CMC.

**Supportive open-set performance evaluation**

The close-set evaluation protocol is fair and reliable for comparing the performances of methods. However, employing an open-set evaluation protocol can be more attractive for practical situations. Therefore, we perform additional experiments using an open-set evaluation protocol. Following [44] with open-set identification experiments, the first 152 subjects (80%) are considered as the training set while the remaining 38 subjects (20%) are considered as the test set. To adapt to the requirements of distinguishing genuine and imposter samples from unseen subjects, we simply train the proposed networks using a Siamese architecture with cosine embedding loss between first and second session's training data. During testing, we compute the cosine similarities between the embeddings from first and second session's test data.

Figure 7(a)/(b) shows the comparative experimental results with the state-of-the-art methods under the feature comparison scenario with the open-set evaluation protocol. Under this more challenging subject independent protocol, our method consistently outperforms FKNet. However, the deep learning approaches are generally suffered from the insufficient training samples for their full capabilities. Therefore, it is not surprising that hand crafted approaches perform slightly better in this challenging experimental settings which the statistical distribution across the two imaging session is large and the amount of available training samples is small.

### 4.2  2D Finger Knuckle Recognition

Once the. Similar to reference [29], we also provide supportive experimental results with 2D finger knuckle recognition to ascertain the effectiveness of the proposed method. The HKPolyU Contactless Finger Knuckle Images Database (Version 1.0) [53] provides 2515 images from 503 subjects, each with five images. We adopt the same challenging evaluation protocol as in reference [29] that, the first major finger knuckle images of each subject are used as the train set while the other four images of each subject are used as the test set. During training, rotationally shifted versions of ±10 degrees of the training images are used for data augmentation. This protocol generates 1,010,024 (503×502×4) imposter comparison scores and 2012 (503×4) genuine comparison scores. We modify the last classification layer of all the networks to an output of 503 classes for this dataset. The provided segmented images are resized to a resolution of 96×108 and cropped with a resolution of 80×96 for testing and 64×80 for training. Therefore, FKNet produces a feature dimension of 8×10×832=66560 and the kernel size of its classification layer and that of our feature extraction layer is modified to be 8×10.

Similar to our experiments with 3D finger knuckle images, we also compare our proposed method with ResNet-50 [45], DenseNet-121 [50], GoogleNet [51], and FKNet [29]. Since ViT-b-16 [52] does not perform well in this experiment, we remove this baseline for enhancing the contrast of the performance comparisons. From our findings in the ablation studies, the input alignment technique is helpful for the generic deep learning methods. Therefore, this technique is also enabled. For our proposed FKNet+, we select a feature dimension of 600 which produces the best performance.

Figure 8(a)/(b) shows the comparative experimental results with the state-of-the-art methods under the classification scenario. It can be observed that our proposed method outperforms other generic deep learning methods and achieves comparable performance to FKNet with a huge reduction of feature dimension size from 66560 to 600. Figure 9(a)/(b) shows the comparative experimental results with the state-of-the-art methods under the feature comparison scenario. Our proposed method achieves slightly better performance in the ROC curves and significantly outperforms all the baselines in the CMC curves.

### 4.3 3D Palmprint Recognition

The scope of this paper is on 3D finger knuckle and available datasets are quite limited. To provide more experimental evidence, we also perform experiments using a 3D palmprint dataset for verifying the generalizability of FKNet+. Following FKNet [29], we use the HKPolyU Contact-free 3D/2D Hand Images Database [54] with exactly same protocol. This database provides two-sessions images from 177 subjects, each with five samples per session. This evaluation protocol uses all first session images for the training and the second session images for the testing, which generates 885 (177×5) genuine and 155760 (177×176×5) imposter comparison scores.

Figure 10(a)/(b) shows the comparative experimental results with the state-of-the-art methods under the feature comparison scenario. Our proposed method significantly outperforms all the baselines in both the ROC and the CMC curves, which verifies the effectiveness of the proposed method.

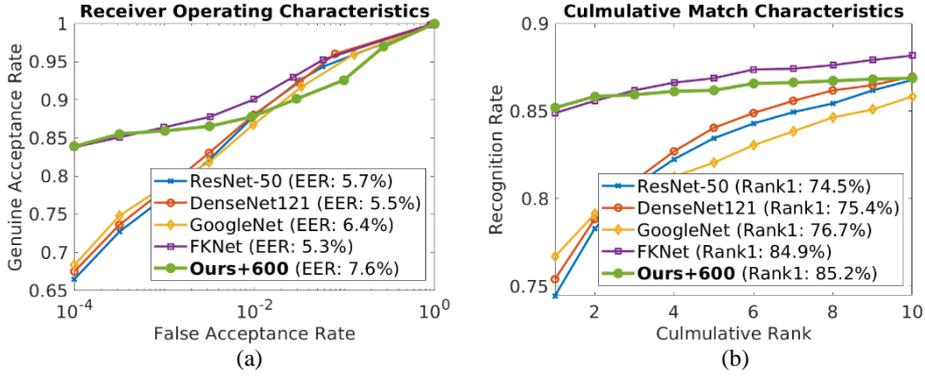

**Fig. 8.** Comparisons using the HKPolyU Contactless Finger Knuckle Images Database [53] under the classification scenario: (a) ROC; (b) CMC.

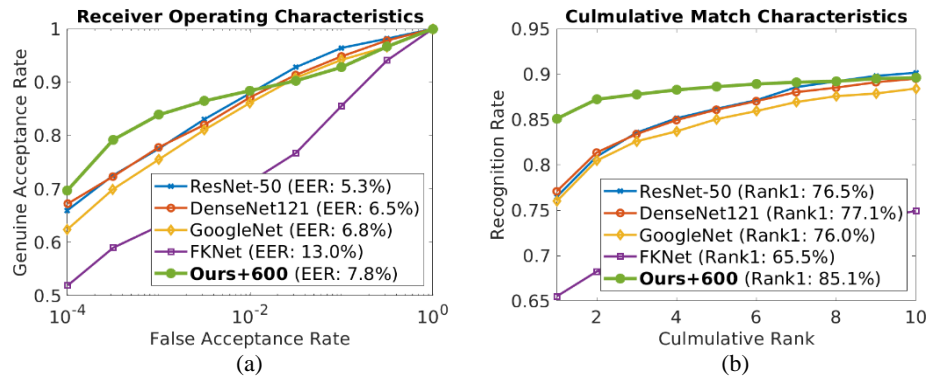

**Fig. 9.** Comparisons using the HKPolyU Contactless Finger Knuckle Images Database [53] under the feature comparison scenario: (a) ROC; (b) CMC.

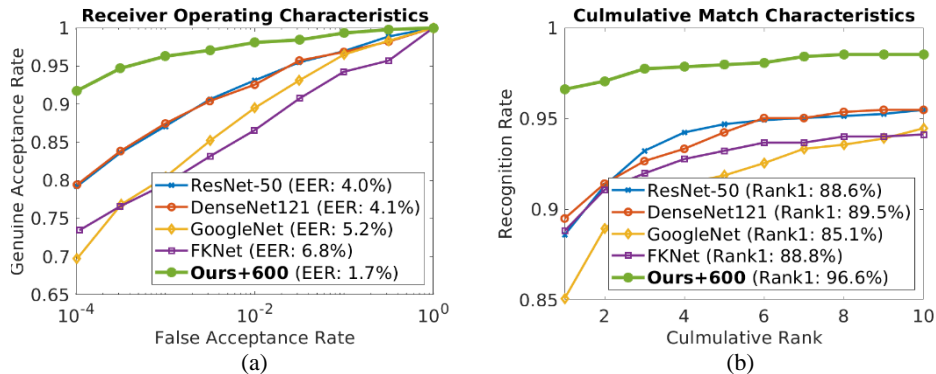

**Fig. 10.** Comparisons using the HKPolyU Contact-free 3D/2D Hand Images Database [54] under the feature comparison scenario: (a) ROC; (b) CMC.

## 5 Conclusions

This paper advances the state-of-the-art 3D finger knuckle recognition method, FKNet [29], by developing an enhanced approach to represent each 3D finger knuckle image using a feature vector with a dimension of only 600. With this simple approach proposed in this paper, we achieve better performances, with a huge reduction in feature size (e.g., from 49920 to 600), which is highly desirable for deploying biometric systems in the real world. In addition, we evaluate the performances using a more practical feature comparison scenario, i.e., comparing the similarities between the extracted features without using the trained classification layer. Our approach outperforms state-of-the-art methods in both verification and identification tasks. Last but not least, this paper provides open-source implementation codes [30] using the current latest version of PyTorch framework, which can help to further advance research on 3D finger knuckle recognition.

Despite the slight performance differences between the proposed deep learning based approach and the hand crafted approaches, this paper advances the high potential deep learning based solution and opens the needs for further developing deep learning based solution for the challenging open set 3D finger knuckle recognition. Incorporating deep metric learning techniques will be a promising future extension. Simplifying both the network architecture and the feature templates will also be important future research directions. We believe that the contributions from this paper can help the development of more robust and useful 3D finger knuckle recognition technology.

**Acknowledgments.** This work was supported by the Academy of Finland for Academy Professor project EmotionAI (grants 336116, 345122) and Infotech Oulu, as well as the CSC-IT Center for Science, Finland, for computational resources.

## References

1. N. Méndez-Llanes, K. Castillo-Rosado, H. Méndez-Vázquez, M. and Tistarelli. On the use of local fixations and quality measures for deep face recognition. IEEE Transactions on Biometrics, Behavior, and Identity Science (TBIOM), 2022.
2. K.H. Cheng, Z. Yu, H. Chen and G. Zhao. Benchmarking 3D Face De-Identification with Preserving Facial Attributes. In IEEE International Conference on Image Processing (ICIP), 2022.
3. K. Wang and A. Kumar. Toward more accurate iris recognition using dilated residual features. IEEE Transactions on Information Forensics and Security (TIFS), 14(12), pp.3233-3245. 2019.
4. M.K. Morampudi, M.V. Prasad and U.S.N. Raju. Privacy-preserving and verifiable multi-instance iris remote authentication using public auditor. Applied Intelligence, 51(10), pp.6823-6836, 2021.
5. R. Ahila Priyadharshini, S. Arivazhagan and M. Arun. A deep learning approach for person identification using ear biometrics. Applied intelligence, 51(4), pp.2161-2172, 2021.
6. F.I. Eyiokur, D. Yaman and H.K. Ekenel. Domain adaptation for ear recognition using deep convolutional neural networks. IET Biometrics, 7(3), pp.199-206, 2018.
7. C. Lin and A. Kumar. Contactless and partial 3D fingerprint recognition using multi-view deep representation. Pattern Recognition (PR), 83, pp.314-327, 2018.
8. X. Yin, Y. Zhu and J. Hu. Contactless fingerprint recognition based on global minutia topology and loose genetic algorithm. IEEE Transactions on Information Forensics and Security (TIFS), 15, pp.28-41, 2019.
9. A. Genovese, V. Piuri, K.N. Plataniotis, and F. Scotti. PalmNet: Gabor-PCA convolutional networks for touchless palmprint recognition. IEEE Transactions on Information Forensics and Security (TIFS), 14(12), pp.3160-3174, 2019.
10. W.M. Matkowski, T. Chai and A.W.K. Kong. Palmprint recognition in uncontrolled and uncooperative environment. IEEE Transactions on Information Forensics and Security (TIFS), 15, pp.1601-1615, 2019.
11. C. Kauba, B. Prommegger, and A. Uhl. Focussing the beam-a new laser illumination based data set providing insights to finger-vein recognition. In IEEE International Conference on Biometrics Theory, Applications and Systems (BTAS), 2018.
12. K. Castillo-Rosado, M. Linortner, A. Uhl, H. Mendez-Vasquez and J. Hernandez-Palancar. Minutiae-based Finger Vein Recognition Evaluated with Fingerprint Comparison Software. In IEEE International Conference of the Biometrics Special Interest Group (BIOSIG), 2020.

13. J. Schuiki, G. Wimmer and A. Uhl. Vulnerability Assessment and Presentation Attack Detection Using a Set of Distinct Finger Vein Recognition Algorithms. In IEEE International Joint Conference on Biometrics (IJCB), 2021.
14. R. Vyas, H. Rahmani, R. Boswell-Challand, P. Angelov, S. Black, and B.M. Williams. Robust End-to-End Hand Identification via Holistic Multi-Unit Knuckle Recognition. In IEEE International Joint Conference on Biometrics (IJCB), 2021.
15. R. Vidhyapriya and R.S. Lovelyn. Personal authentication mechanism based on finger knuckle print. Journal of Medical Systems, 43(8), pp.1-7, 2019.
16. A. Muthukumar and A. Kavipriya. A biometric system based on Gabor feature extraction with SVM classifier for Finger-Knuckle-Print. Pattern Recognition Letters (PRL), 125, pp.150-156, 2019.
17. G. Jaswal and R.C. Poonia. Selection of optimized features for fusion of palm print and finger knuckle-based person authentication. Expert Systems, 38(1), p.e12523, 2021.
18. S. Li, B. Zhang, L. Fei, and S. Zhao. Joint discriminative feature learning for multimodal finger recognition. Pattern Recognition (PR), 111, p.107704, 2021.
19. N.E. Chalabi, A. Attia, and A. Bouziane. Multimodal finger dorsal knuckle major and minor print recognition system based on PCANET deep learning. ICTACT J Image Video Process, 10(3), pp.2153-2158, 2020.
20. F. Liu and X. Liu. Voxel-based 3D Detection and Reconstruction of Multiple Objects from a Single Image. In Advances in Neural Information Processing Systems (NIPS), 34, 2021.
21. K.H. Cheng and A. Kumar. Revisiting outlier rejection approach for non-Lambertian photometric stereo. IEEE Transactions on Image Processing (TIP), 28(3), pp.1544-1555, 2018.
22. F. Liu, Q. Zhao, X. Liu, and D. Zeng. Joint face alignment and 3D face reconstruction with application to face recognition. IEEE Transactions on Pattern Analysis and Machine Intelligence (TPAMI), 42(3), pp.664-678, 2018.
23. Z. Zhang, F. Da, and Y. Yu. Learning directly from synthetic point clouds for "in-the-wild" 3D face recognition. Pattern Recognition (PR), 123, p.108394, 2022.
24. D. L. Woodard and P. J. Flynn. Finger Surface as a Biometric Identifier. Computer Vision and Image Understanding (CVIU), 100(3), pp. 357-384, 2005.
25. K.H. Cheng and A. Kumar. Advancing surface feature encoding and matching for more accurate 3D biometric recognition. In IEEE International Conference on Pattern Recognition (ICPR), 2018.
26. K.H. Cheng and A. Kumar. Contactless biometric identification using 3D finger knuckle patterns. IEEE Transactions on Pattern Analysis and Machine Intelligence (TPAMI), 42(8), pp.1868-1883, 2019.
27. K. Sundararajan and D.L. Woodard. Deep learning for biometrics: A survey. ACM Computing Surveys (CSUR), 51(3), pp.1-34, 2018.
28. M. Vatsa, R. Singh and A. Majumdar. Deep learning in biometrics. CRC Press. 2018.
29. K.H. Cheng and A. Kumar. Deep Feature Collaboration for Challenging 3D Finger Knuckle Identification. IEEE Transactions on Information Forensics and Security (TIFS), 16, pp.1158-1173, 2020.
30. Weblink for downloading our implementation codes with PyTorch 1.11.0. https://github.com/kevinhmcheng/deep-3d-finger-knuckle2.
31. G. Jaswal, A. Kaul, and R. Nath. Knuckle Print Biometrics and Fusion Schemes – Overview, Challenges, and Solutions. ACM Computing Surveys (CSUR), 49(2), pp. 1-46, 2016.
32. H.M. Cheng. Contactless 3D finger knuckle identification. (https://theses.lib.polyu.edu.hk/handle/200/11223), 2021.
33. L. Zhang, L. Zhang, D. Zhang and H. Zhu. Online Finger Knuckle-Print Verification for Personal Authentication. Pattern Recognition (PR), 43(7), pp. 2560-2571, 2010.
34. Z.S. Shariatmadar and K. Faez. A novel approach for Finger-Knuckle-Print recognition based on Gabor feature fusion. In International Congress on Image and Signal Processing, 2011.
35. L. Zhang, H. Li, and Y. Shen. A novel Riesz transforms based coding scheme for finger-knuckle-print recognition. In International Conference on Hand-based Biometrics, 2011.
36. R. Vyas, H. Rahmani, R. Boswell-Challand, P. Angelov, S. Black, and B.M. Williams. Robust End-to-End Hand Identification via Holistic Multi-Unit Knuckle Recognition. In IEEE International Joint Conference on Biometrics (IJCB), 2021.
37. S. Lakshmanan, P. Velliyan, A. Attia and N.E. Chalabi. Finger knuckle pattern person authentication system based on monogenic and LPQ features. Pattern Analysis and Applications, 25(2), pp.395-407, 2022.
38. A.S. Tarawneh, A.B. Hassanat, E.A. Alkafaween, B. Sarayrah, S. Mnasri, G.A. Altarawneh, M. Alrashidi, M. Alghamdi and A. Almuhaimeed. DeepKnuckle: Deep Learning for Finger Knuckle Print Recognition. Electronics, 11(4), p.513, 2022.
39. M.C. Younis and H. Abuhammad. A hybrid fusion framework to multi-modal biometric identification. Multimedia Tools and Applications, 80(17), pp.25799-25822, 2021.
40. J. Khodadoust, M.A. Medina-Pérez, R. Monroy, A.M. Khodadoust, and S.S. Mirkamali. A multibiometric system based on the fusion of fingerprint, finger-vein, and finger-knuckle-print. Expert Systems with Applications, 176, p.114687, 2021.
41. M. Chaa, Z. Akhtar and A. Lati. Contactless person recognition using 2D and 3D finger knuckle patterns. Multimedia Tools and Applications, 81(6), pp.8671-8689, 2022.
42. K.H. Cheng and A. Kumar. Efficient and Accurate 3D Finger Knuckle Matching Using Surface Key Points. IEEE Transactions on Image Processing (TIP), 29, pp.8903-8915, 2020.
43. K.H. Cheng and A. Kumar. Distinctive Feature Representation for Contactless 3D Hand Biometrics using Surface Normal Directions. In IEEE International Joint Conference on Biometrics (IJCB), 2020.

44. K.H. Cheng and A. Kumar. Accurate 3D Finger Knuckle Recognition Using Auto-Generated Similarity Functions. IEEE Transactions on Biometrics, Behavior, and Identity Science (TBIOM), 3(2), pp.203-213, 2021.
45. K. He, X. Zhang, S. Ren, and J. Sun. Deep residual learning for image recognition. In IEEE Conference on Computer Vision and Pattern Recognition (CVPR), 2016.
46. K. He, G. Gkioxari, P. Dollár, and R. Girshick. Mask r-cnn. In IEEE International Conference on Computer Vision (ICCV), 2017.
47. T.Y. Lin, M. Maire, S. Belongie, J. Hays, P. Perona, D. Ramanan, P. Dollár, and C.L. Zitnick. Microsoft coco: Common objects in context. In European Conference on Computer Vision (ECCV), 2014.
48. R.J. Woodham. Photometric method for determining surface orientation from multiple images. Optical engineering, 19(1), pp.139-144, 1980.
49. The HKPolyU 3D Finger Knuckle Images Database: https://www4.comp.polyu.edu.hk/~csajaykr/3DKnuckle.htm
50. G. Huang, Z. Liu, L.V.D. Maaten, and K.Q. Weinberger. Densely connected convolutional networks. In IEEE Conference on Computer Vision and Pattern Recognition (CVPR), 2017.
51. C. Szegedy, W. Liu, Y. Jia, P. Sermanet, S. Reed, D. Anguelov, D. Erhan, V. Vanhoucke, and A. Rabinovich. Going deeper with convolutions. In IEEE Conference on Computer Vision and Pattern Recognition (CVPR), 2015.
52. A. Dosovitskiy, L. Beyer, A. Kolesnikov, D. Weissenborn, X. Zhai, T. Unterthiner, M. Dehghani, M. Minderer, G. Heigold, S. Gelly and J. Uszkoreit. An image is worth 16x16 words: Transformers for image recognition at scale. arXiv preprint arXiv:2010.11929, 2020.
53. The HKPolyU Contactless Finger Knuckle Images Database (V-1.0): http://www4.comp.polyu.edu.hk/~csajaykr/fn1.htm
54. V. Kanhangad, A. Kumar, and D. Zhang. A Unified Framework for Contactless Hand Verification. IEEE Trans. Info. Foren. & Sec., 2011.